\documentclass[runningheads]{llncs}

\usepackage{eccv}

\usepackage{eccvabbrv}

\usepackage{graphicx}
\usepackage{booktabs}
\usepackage{adjustbox}
\usepackage[numbers]{natbib}
\usepackage{cuted}  
\usepackage{capt-of}  
\usepackage{dblfloatfix}
\usepackage{wrapfig}
\usepackage[table]{xcolor}
\usepackage{colortbl}
\graphicspath{{./image/}}
\usepackage{algorithm}
\usepackage{algorithmicx}
\usepackage[section]{placeins}
\usepackage{amssymb}
\usepackage{float}
\usepackage{color, soul}
\usepackage{enumitem}
\usepackage{graphicx}
\usepackage{subcaption}
\usepackage[T1]{fontenc}
\usepackage{booktabs}
\usepackage{multicol}
\usepackage{subcaption}
\usepackage{multirow}
\usepackage{dirtytalk}

\usepackage{amsmath}
\usepackage{makecell}
\usepackage{cuted}
\usepackage{capt-of}
\usepackage{amsmath} 
\usepackage[table,xcdraw]{xcolor}
\usepackage{multirow}
\usepackage{amssymb}
\usepackage{float}
\usepackage{color, soul}
\usepackage{enumitem}
\usepackage{graphicx}
\usepackage{subcaption}
\usepackage[T1]{fontenc}
\usepackage{booktabs}
\usepackage{multicol}
\usepackage{subcaption}
\usepackage{multirow}
\usepackage{dirtytalk}
\usepackage{multirow}
\usepackage[table,xcdraw]{xcolor}
\usepackage[utf8]{inputenc}
\usepackage{xcolor} 

\usepackage{adjustbox}
\usepackage{graphicx}
\usepackage{subcaption}
\usepackage{adjustbox}
\usepackage{pgffor}
\usepackage{graphicx}
\usepackage{flafter}
\usepackage[section]{placeins}
\usepackage{makecell}
\usepackage{currfile}
\usepackage{ulem}
\usepackage{tabularx}
\usepackage{xcolor}
\usepackage{algorithm}
\usepackage{algorithmicx}
\usepackage{algpseudocode}
\usepackage{amsmath}
\newcommand{\our}{UNITY}
\usepackage{caption}

\usepackage[accsupp]{axessibility}  

\usepackage{hyperref}

\usepackage{orcidlink}

\begin{document}

\title{UNITY: Attention Flow Networks for Adaptive Conditioning in Diffusion}

\titlerunning{UNITY}

\author{
Aryan Das\inst{1}\orcidlink{0000-0003-0439-7351} \and
Koushik Biswas\inst{2}\orcidlink{0000-0002-9818-8966} \and
Moloud Abdar\inst{3}\orcidlink{0000-0002-3059-6357} \and
Vinay Kumar Verma\inst{4}\orcidlink{0000-0003-0172-402X}
}

\authorrunning{A. Das et al.}


\institute{
VIT Bhopal, India \and
IIIT Delhi, India \and
The University of Queensland, Australia \and
IIT Kanpur, India\\
\email{
$^{1}$~aryan.das2021@vitbhopal.ac.in,
$^{2}$~koushikb@iiitd.ac.in, \\
\{$^{3}$~m.abdar1987, $^{4}$~vinayugc\}@gmail.com
}
}

\maketitle
\begin{abstract}
We introduce \our, Universal-to-Specialized adapter for efficient and scalable composite conditioning in diffusion-based image generation. Unlike prior methods that train separate adapters for each condition, \our{} jointly learns shared semantics across multiple conditioning modalities and later specializes without altering the architecture. The proposed two-stage training paradigm consists of a \textit{Universal Stage}, which captures cross-modal joint representations across all conditioning types using half of the total training steps, followed by a \textit{Specialization Stage} that refines modality-specific features with the remaining steps. The core of our design are the \textbf{Morphable Attention Flow (MAF) Network} and \textbf{Morph Wrapper} modules, which enable channel-aware and spatially adaptive feature alignment through learnable flow fields and attention fusion. This constant-complexity formulation supports flexible operation under single or composite conditioning while significantly reducing inference latency and memory footprint. Extensive experiments on multiple datasets demonstrate that \our{} achieves State-of-the-Art (SoTA) image fidelity and memory efficiency. \textbf{Code:} \url{https://github.com/arya-domain/UNITY}
\end{abstract}

\section{Introduction}
\label{sec:intro}

\begin{figure}
    \centering
    \includegraphics[width=\linewidth]{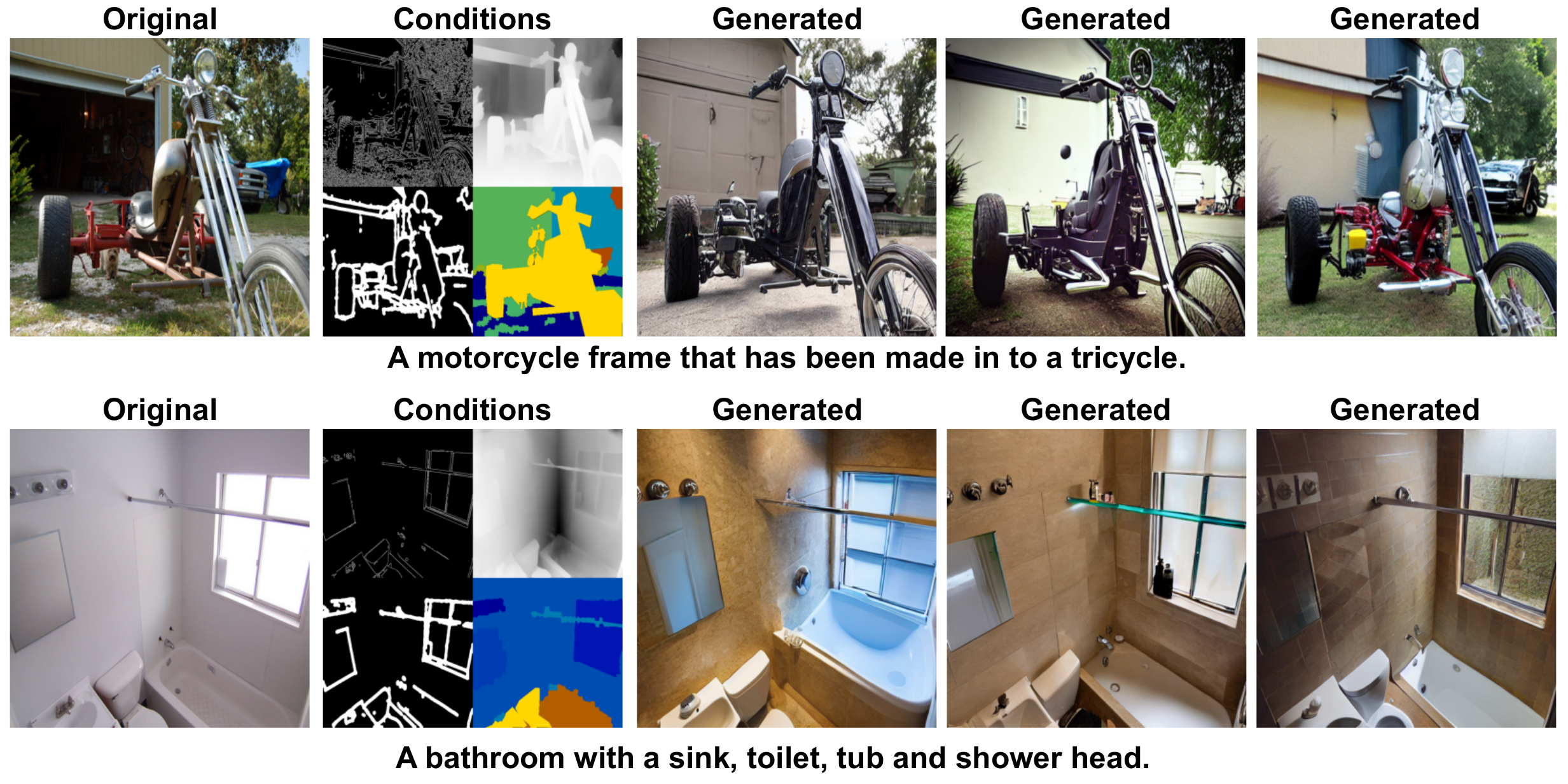}
    \caption{Composite conditioned image generation with \our{}-Adapter. Each example is jointly conditioned on four complementary control signals shown in the \textbf{Conditions} column: \emph{top left}, Canny edge; \emph{top right}, Depth map; \emph{bottom left}, Sketch; and \emph{bottom right}, Semantic Segmentation.}
    \label{fig:placeholder}
\end{figure}

Conditional image generation has become a foundational capability in modern visual AI, yet as real-world applications increasingly demand simultaneous control over structure, semantics, and style, the scalability of existing conditioning frameworks has emerged as a critical and unsolved bottleneck. 
Diffusion-based models~\cite{diff,stablediffusion,ldm}  have enabled high-fidelity synthesis guided by diverse signals such as text, depth, 
sketches, and semantic masks, and attention-based architectures with vision transformers have further strengthened cross-modal alignment~\cite{vit,plinvit,clip,karim2026focus,nexus2026}.
Multi-modal conditioning is not a luxury, it is a 
\begin{wrapfigure}{r}{0.55\linewidth}
    \centering
    \includegraphics[width=\linewidth]{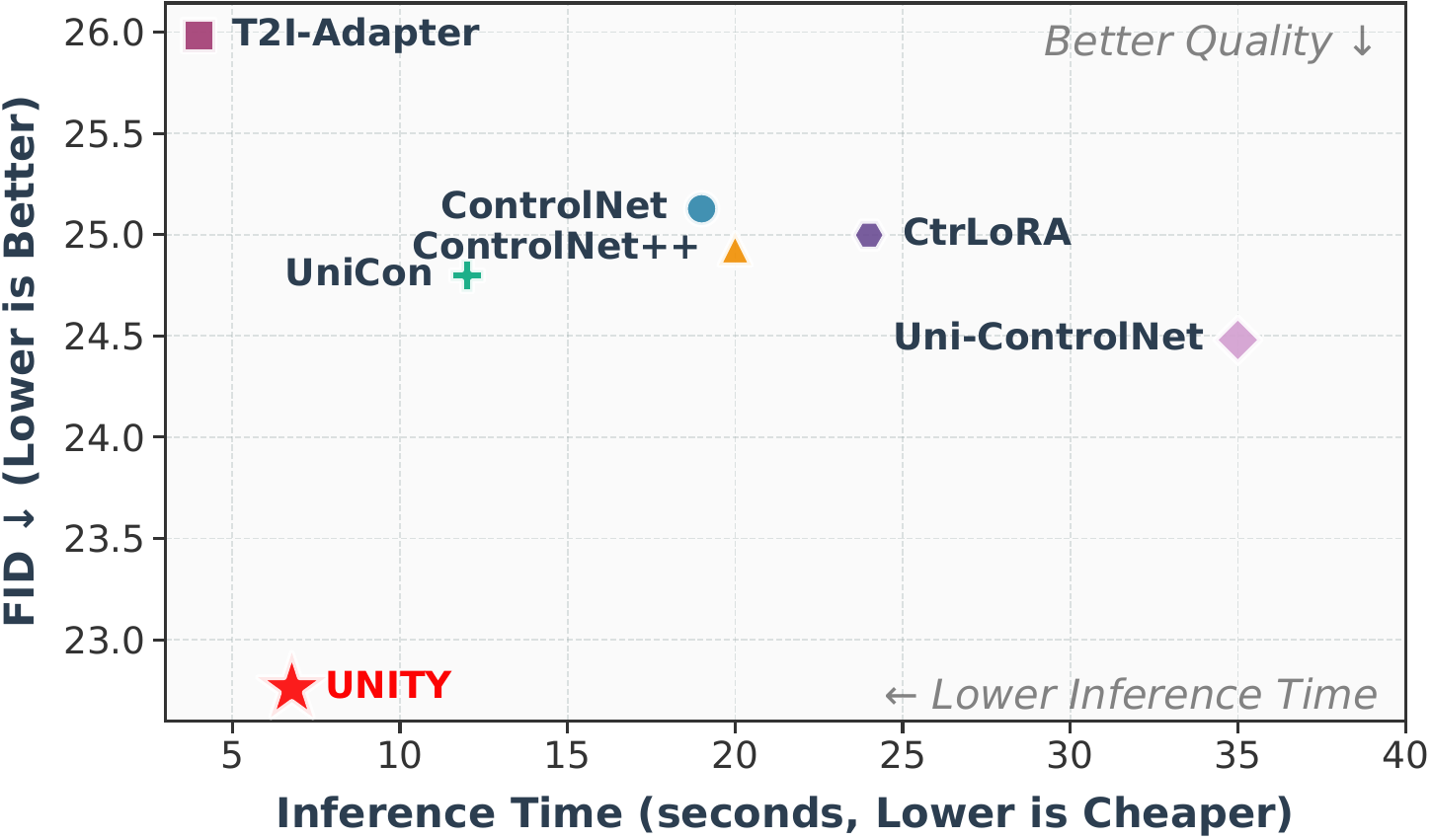}
    \caption{Comparison of FIDs and Infer time across methods.}
    \label{fig:comparison}
\end{wrapfigure}
necessity for structure-preserving tasks such as text-to-image generation~\cite{t2i1,t2i2,t2i5} and image-to-image translation~\cite{i2i2,i2i3}, where spatial layout and semantic consistency are non-negotiable.
Composite conditioning, where multiple inputs jointly describe a shared underlying scene, is the natural evolution of this paradigm, offering richer, more coherent guidance than any single modality can provide. Yet the field has treated this as an afterthought, bolting together independent adapters rather than designing systems that fundamentally understand multi-modal structure. This raises an unavoidable question: \textit{\textbf{Can we design a single adapter that jointly learns shared semantics across all conditioning modalities, scales to any number of conditions without additional cost, and maintains precise spatial alignment?}}

Dominant adapter frameworks, including ControlNet~\cite{controlnet}, T2I-Adapter~\cite{t2i}, ControlNet++~\cite{controlnet++}, and CtrLoRA~\cite{ctrlora}, treat each conditioning modality as an isolated problem, each requiring a fully independent training cycle to reach optimal performance. The consequence is structural and severe: training cost and memory scale linearly with the number of conditions, making four-modality conditioning demand four times the resources of a single-modality baseline. Uni-ControlNet~\cite{unicontrolnet}, the closest attempt 
at unification, still requires $8\times$ the GPU memory of a single-condition setup, with 1271.42M parameters and no mechanism for shared semantic learning across conditions. More fundamentally, none of these architectures possesses a unified mechanism to model fine-grained spatial correspondences across modalities. ControlNet-style adapters duplicate the entire Stable Diffusion encoder, an expensive design choice that paradoxically sacrifices modality-specific adaptability for scale, and every existing method executes a parallel denoising path alongside the diffusion backbone, which is the primary driver of inference latency.

\textit{We argue that this demands not incremental improvement but a ground-up rethinking of how adapters should handle multiple conditioning modalities.} Two core problems must be solved simultaneously, and prior work has ignored both. \textbf{First}, efficient multi-modal conditioning requires a framework that jointly learns the shared semantic structure underlying all modalities, rather than treating each in isolation, and does so without any increase in parameter count or training budget as modalities grow. \textbf{Second}, spatial correspondence across conditioning modalities cannot be captured by attention alone; it demands explicit, learnable geometric alignment that adapts to each modality's spatial characteristics. We introduce \textbf{\our-Adapter}, a Universal-to-Specialized framework that confronts both problems directly. \our-Adapter trains in two decisive stages: a \textit{Universal Stage} that jointly learns from all $K$ conditions in the first $N/2$ steps, forging a shared discriminative representation space, followed by a \textit{Specialization Stage} that refines each modality individually in the remaining $N/2$ steps, delivering a provable $37.5\%$ reduction in training cost for four conditions with \textit{strictly fixed} parameters regardless of $K$. Complementing this, the proposed \textbf{Morphable Attention Flow (MAF)} network introduces learnable displacement fields that directly capture spatial correspondences between conditioning inputs and latent representations, enabling precise geometric alignment without backbone duplication and without any parallel denoising path, substantially cutting inference latency. The result is striking: \our-Adapter achieves State-of-the-Art (SoTA) FID and CLIP scores across all four conditioning modalities on a single 24GB GPU, outperforming methods that require $4\times$ to $8\times$ more memory, as shown in Figure~\ref{fig:comparison}.
Our contributions are summarized as follows:
\begin{itemize}[leftmargin=*]
    \item We propose \textbf{\our-Adapter}, a universal-to-specialized framework that enables efficient and flexible composite conditioning across multiple modalities without added model complexity or cost.
    \item We introduce the \textbf{Morphable Attention Flow (MAF)} network with \textbf{Morph Wrapper} modules, which performs channel-aware and spatially adaptive feature alignment using learnable flow fields and attention-based fusion without duplicating the diffusion backbone.
\end{itemize}
The extensive experiments over the various datasets demonstrate that it achieves the SoTA fidelity and memory efficiency while significantly reducing memory and inference overhead compared to existing multi-adapter frameworks.
\section{Related Work}
\label{sec:related}

\noindent\textbf{Conditional Diffusion Models:}
Text-to-image diffusion models have achieved remarkable success in high-fidelity image synthesis~\cite{diff,ldm,stablediffusion,yu2025ttsnap,t2i1}. Controllable generation frameworks introduce additional conditioning modalities beyond text. ControlNet~\cite{controlnet} pioneered trainable adapters via UNet encoder duplication with zero convolution initialization. T2I-Adapter~\cite{t2i} proposed lightweight adapters with faster inference. Uni-ControlNet~\cite{unicontrolnet} unified multiple controls in all-in-one architectures. UniCon~\cite{unicon} learned joint image-condition distributions enabling flexible sampling. ControlNet++~\cite{controlnet++} improved consistency through iterative feedback. UniControl~\cite{UniControl} extended multi-modal controllability via task-specific training. CtrLoRA~\cite{ctrlora} explored LoRA-based extensible control. Despite effectiveness, these require independent training per condition, causing linear scaling in cost and memory. They perform parallel denoising, increasing inference latency. Our framework maintains constant complexity via the training paradigm, enabling global guidance without architectural duplication.

\noindent\textbf{Multi-Modal and Composite Conditioning:}
Handling multiple conditioning inputs remains a core challenge in controllable generation~\cite{UAVLS2026}. Composer~\cite{composer} introduced composable conditions via a Product-of-Experts formulation, and multi-modal GANs~\cite{mul1} studied composite guidance in adversarial settings. Transformer based methods have pursued unified control: UniCombine~\cite{unicombine} combined condition tokens for multi-conditional generation in DiT models, and OminiControl~\cite{ominicontrol} achieved unified sequence processing with minimal parameter overhead. However, these methods either lack efficient training for composite scenarios or require separate adaptation for each condition set. TCIG~\cite{tcig} used a two-stage framework that separates controllability from quality enhancement but emphasizes sequential refinement rather than joint multi-modal learning. In contrast, \our-Adapter learns shared semantic representations across all conditions in a universal stage before specializing to individual modalities, enabling flexible composite control without retraining.

\noindent\textbf{Flow-Based Feature Alignment:}
Flow networks have shown strong efficacy for spatial feature transformation and alignment in generative tasks. Normalizing flows~\cite{flow} introduced invertible transformations for density modeling, later extended by Diffusion Normalizing Flow~\cite{diffnorm}, which integrates Stochastic Diffusion~\cite{song2021scorebased} with learnable flow dynamics. In conditional generation, optical flow-based warping has been widely adopted: Global-Flow Local-Attention~\cite{gfla} enables feature-level spatial reassembly for pose-guided synthesis, Latent Flow Diffusion Models~\cite{lfdm} generate latent flow sequences for video, and Flow-based Feature Warping~\cite{ffwm} achieves illumination preserving frontalization via learned flow fields. These works demonstrate that learnable flows support spatial alignment across modalities. Building on this insight, our MAF introduces channel-aware, attention-driven warping within Morph Wrapper modules. Unlike optical flow, which models only spatial displacement, MAF learns parametric flow fields that adaptively align multi-modal conditioning features with denoising features, enabling semantically consistent fusion while preserving spatial structure.

\noindent\textbf{Training Strategies for Controllable Diffusion:}
Efficient training paradigms are crucial for scaling diffusion models. Adapter-based methods~\cite{adapter,vitadapter} offer parameter efficient transfer learning in vision transformers, yet their use for composite conditioning remains limited. Two-stage strategies such as Diff2Flow~\cite{diff2flow} bridge diffusion and flow matching via timestep rescaling and velocity alignment but focus on adaptation rather than control. 
Our \our{}-Adapter framework introduces a training paradigm tailored to composite conditioning: a universal stage learns shared cross-modal representations, followed by a specialization stage that refines individual condition handling. This design keeps total training cost constant regardless of the number of conditions, unlike prior methods that require separate training cycles per condition.
\section{Methodology}

\begin{figure*}[t]
    \centering
    \includegraphics[width=1\linewidth]{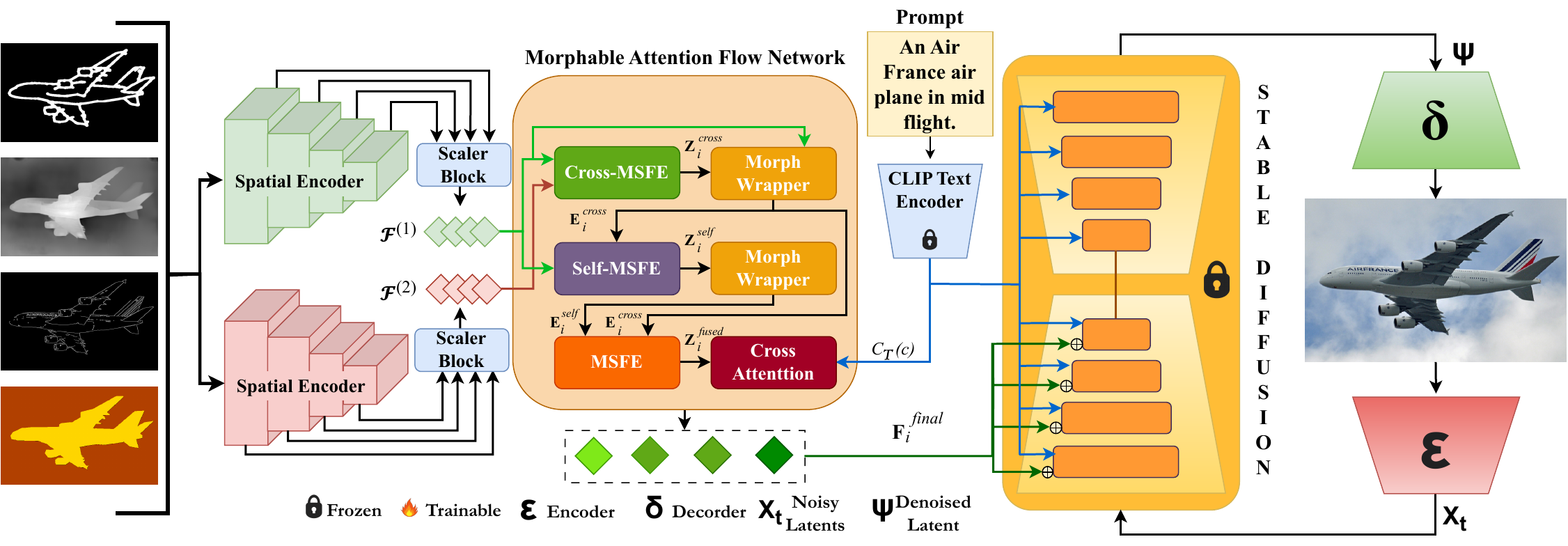}
    \caption{Proposed Framework: Conditioning inputs are processed through dual Spatial Encoders with Scaler Block to produce multi-scale feature pyramids $\mathcal{F}^{(1)}$ and $\mathcal{F}^{(2)}$. The Morphable Attention Flow (MAF) Network performs Cross-Modal alignment (Cross-MSFE + Morph Wrapper), Self-Refinement (Self-MSFE + Morph Wrapper), and Fusion (MSFE), with final features $\mathbf{F}_i^{final}$ conditioned via Cross-Attention with Text Embeddings $\mathcal{C}_T(c)$ and injected into the frozen Stable Diffusion's UNet and fused with Point-Wise Addition.}
    \label{fig:model}
\end{figure*}

\subsection{Preliminaries}
Diffusion models generate samples by reversing a gradual noising process. Given a clean image \(\mathbf{X}_{0}\), the forward diffusion process constructs progressively noisier states according to \(q(\mathbf{X}_{t} \mid \mathbf{X}_{t-1}) = \mathcal{N}(\mathbf{X}_{t}; \sqrt{1-\beta_{t}}\,\mathbf{X}_{t-1},\,\beta_{t}\mathbf{I})\), where \(\{\beta_{t}\}_{t=1}^{T}\) is a predefined noise schedule that produces approximately Gaussian noise after \(T\) steps. The reverse process aims to denoise by modeling \(p_{\theta}(\mathbf{X}_{t-1} \mid \mathbf{X}_{t}) = \mathcal{N}(\mathbf{X}_{t-1}; \mu_{\theta}(\mathbf{X}_{t}, t), \Sigma_{\theta}(t))\), typically reparameterized using a noise prediction network \(\epsilon_{\theta}\) trained via the objective \(\mathcal{L}(\theta) = \mathbb{E}_{t,\mathbf{X}_{0},\epsilon}[\|\epsilon - \epsilon_{\theta}(\mathbf{X}_{t}, t)\|^{2}]\). In this work, we employ a pretrained Latent Diffusion Model (LDM) based on the Stable Diffusion~\cite{stablediffusion} architecture, which operates in a compressed latent space using a variational autoencoder (VAE) consisting of an encoder \(\varepsilon\), a decoder \(\delta\), and a denoising UNet. The encoder maps an input image \(\mathbf{X}_{0}\) to its latent representation \(\Psi = \varepsilon(\mathbf{X}_{0})\), while the decoder reconstructs the image from the latent as \(\hat{\mathbf{X}} = \delta(\Psi)\). Forward diffusion is then applied in the latent space, perturbing \(\Psi\) over \(T\) timesteps to produce noisy latents \(\mathbf{X}_{t}\). The UNet, parameterized by \(\theta\), performs reverse denoising according to \(\mathbf{X}_{t-1} = f_{\theta}(\mathbf{X}_{t}, t, c)\), where \(c\) denotes the text-conditioning vector obtained from a frozen CLIP text encoder \(\mathcal{C}_{T}\). The conditioning is integrated into the UNet via cross-attention mechanisms at each denoising step.

\subsection{Framework}
\label{sec:framework}

\our{} operates as a Plug-and-Play module for the frozen SD1.5 UNet backbone and text-encoder $\mathcal{C_T}$. The core innovation is our Flow-based Adapter design (Fig.~\ref{fig:model}) coupled with the two-stage training paradigm in Algorithm~\ref{algo:traning} that yields $\frac{K-1}{2K}$ training reduction (37.5\% for $K=4$). Rather than parallel condition-specific denoising, \our-Adapter provides global multi-scale guidance to the frozen SD1.5, which performs core denoising while the adapter ensures conditional alignment across timesteps. The adapter maintains constant parameters regardless of $K$, enabling flexible single or composite conditioning at inference. Text embeddings $\mathcal{C_T}$ maintain semantic coherence between visual conditions and textual intent, enhancing structural consistency without compromising generative quality or computational efficiency.

\subsection{Adapter Architecture}

\our{} introduces a unified conditioning mechanism for the frozen Stable Diffusion UNet through two key components: \textit{Spatial Encodings} and the \textit{Morphable Attention Flow (MAF) Network}, which consists of \textit{Multi-Scale Flow Estimators} (MSFE) and \textit{Morph Wrappers}. The spatial encoders transform conditioning inputs into hierarchical feature pyramids that are modulated by a Scaler Block and aligned with the UNet's multi-scale architecture. The MAF Network then processes these features through three sequential stages: \textit{Cross-Modal Alignment}, where a Cross-MSFE followed by a Morph Wrapper establishes geometric correspondences between complementary conditioning features; \textit{Self-Refinement}, where a Self-MSFE and Morph Wrapper further enhance intra-modal structural consistency; and \textit{Fusion}, where a refinement MSFE integrates the cross-aligned and self-refined representations into unified multi-scale embeddings. Finally, the fused features are conditioned through Cross-Attention with the text embeddings $\mathcal{C}_{T}(c)$ before being injected into the frozen Stable Diffusion UNet via point-wise addition.

\subsubsection{Spatial Encoding}

The spatial encoding module transforms conditional inputs into hierarchical feature representations aligned with the UNet’s multi-scale architecture. 
It comprises two primary components: a \textit{Spatial Encoder}, which extracts multi-scale features, and a \textit{Scaler Block}, which adaptively modulates these features before integration into the UNet.

\noindent
\textbf{Spatial Encoder:}
The spatial encoder constructs a hierarchical feature pyramid that preserves spatial fidelity while aligning with the multi-scale UNet structure. Pixel unshuffle performs a lossless spatial-to-channel transformation, and pre-activation residual blocks ensure stable gradient propagation.
Given $K$ conditioning inputs $\mathbf{C} \in \mathbb{R}^{H \times W \times C_k}$ with $C_k = 3K$, we first apply pixel unshuffle with scale factor $s$:
\begin{equation}
\mathbf{F}_0 = \mathit{PixelUnshuffle}(\mathbf{C}),
\quad
\mathbf{F}_0 \in \mathbb{R}^{\frac{H}{s} \times \frac{W}{s} \times C_k s^2}.
\end{equation}
The encoder then produces multi-level features $\{\mathbf{F}_i\}_{i=1}^{L}$ through $L$ hierarchical stages:
\begin{equation}
\mathbf{F}_i =
\mathcal{R}_i^{(2)} \big(
\mathcal{R}_i^{(1)}(
\mathcal{D}_i(\mathbf{F}_{i-1})
)\big),
\quad i = 1,\ldots,L.
\end{equation}
Here, $\mathcal{D}_i$ denotes a strided $3\times3$ convolution that reduces spatial resolution and increases channel dimension from $d_{i-1}$ to $d_i$.  
Each $\mathcal{R}_i^{(j)}$ is a pre-activation residual block defined as, 
$\mathcal{R}(\mathbf{X}) = \mathbf{X} + \mathcal{G}(\mathbf{X}),$
where $\mathcal{G}(\cdot)$ consists of two $3\times3$ convolutions with BN and ReLU in pre-activation order.
All convolutions use padding to preserve spatial resolution, and bias terms are omitted due to normalization.  
This formulation yields a computationally efficient multi-scale encoder while retaining dense spatial information.

\noindent
\textbf{Scaler Block:}
The Scaler Block applies adaptive channel-wise modulation to encoded features before fusion. This adaptive modulation enables coherent feature fusion across heterogeneous conditioning types while preserving spatial fidelity.
For each encoder output $\mathbf{F}_{i} \in \mathbb{R}^{H_{i} \times W_{i} \times d_{i}}$, the block performs a learnable $1\times1$ transformation:
\begin{equation}
\tilde{\mathbf{F}}_{i} 
= 
\mathit{Conv}^{1 \times 1}_{d_{i} \rightarrow d_{i}}(\mathbf{F}_{i}),
\quad 
i \in \{1, \ldots, L\}.
\end{equation}
These pointwise convolutions act as gating mechanisms that emphasize or suppress channels based on the conditioning modality, enabling dynamic feature reweighting while maintaining spatial structure.  
The set of modulated features $\{\tilde{\mathbf{F}}_{i}\}_{i=1}^{L}$ is further processed through the MAF Network.

\subsubsection{Morphable Attention Flow Network}

\begin{figure*}[t]
\centering
\begin{minipage}[t]{0.47\textwidth}
\vspace{0pt}
\centering
\hrule
\captionsetup{type=algorithm}
\captionof{algorithm}{UNITY: Two-Stage Training}
\label{algo:training}
\hrule
\begin{algorithmic}[1]
\State \textbf{Input:} $\{C_i\}_{i=1}^K$, $N$, frozen UNet $\theta$
\State \textbf{Universal Stage ($N/2$ steps)}
\State $\phi \gets \mathrm{Pretrain}(\{C_i\}, N/2)$
\State \textbf{Specialization Stage ($N/2$ steps)}
\For{$i=1$ to $K$}
    \State $\phi_i \gets \mathrm{Finetune}(C_i, N/2)$
\EndFor
\State $\phi \gets \bigcup_i \phi_i$
\State $T_{\our}=\frac{N}{2}(1+K)$ vs.\ $KN$
\end{algorithmic}
\hrule
\end{minipage}
\hfill
\begin{minipage}[t]{0.51\textwidth}
\vspace{0pt}
\centering
\includegraphics[width=\linewidth]{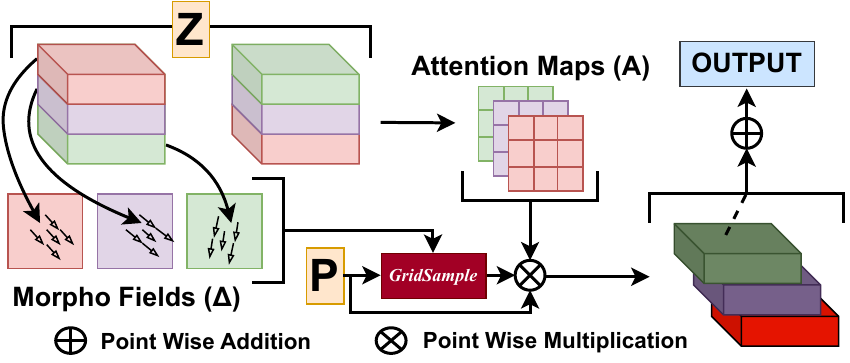}
\captionof{figure}{Morph Wrapper: Adaptive feature warping using learned Morpho Fields $\mathbf{\Delta}$ and Attention Weighted Aggregation across $M$ sampling points.}
\label{fig:wrap}
\end{minipage}

\end{figure*}

The \textit{Morphable Attention Flow (MAF) Network} constitutes the core processing that transforms spatial embeddings through iterative refinement via flow estimation and morphological wrapping. Operating on feature pyramids from dual spatial encoders, the MAF Network performs cross-modal alignment and self-refinement across $L$ hierarchical scales.
Two parallel spatial encoding pathways process complementary conditioning signals, each producing multi-scale feature pyramids through the spatial encoding module described previously. The outputs from the first and second spatial encoders, after their respective \textit{Scaler Blocks}, yield two feature sets $\mathcal{F}^{(1)} = \{\tilde{\mathbf{F}}^{(1)}_{i}\}_{i=1}^{L}$ and $\mathcal{F}^{(2)} = \{\tilde{\mathbf{F}}^{(2)}_{i}\}_{i=1}^{L}$, where $\tilde{\mathbf{F}}^{(1)}_{i}, \tilde{\mathbf{F}}^{(2)}_{i} \in \mathbb{R}^{C_{i} \times H_{i} \times W_{i}}$ represent modulated features at scale $i$. These dual representations capture complementary structural and semantic information that will be progressively refined through the MAF Network's stage processing design.
For each scale $i \in \{1, \ldots, L\}$, processing begins with cross-modal flow estimation, where features from both encoders are concatenated and fed into a MSFE (Section~\ref{sec:msfe}) to establish inter-modal correspondences: 
\begin{equation}
\mathbf{Z}^{cross}_{i} = \textit{MSFE}_{cross}\Big(\text{Concat}\big(\tilde{\mathbf{F}}^{(1)}_{i}, \tilde{\mathbf{F}}^{(2)}_{i}\big)\Big),
\end{equation}
where $\mathbf{Z}^{cross}_{i}$ encodes cross-modal flow features. This $MSFE_{cross}$ output captures the geometric and semantic relationships between the two encoding features, providing the foundation for subsequent morphological alignment.
The cross-modal features $\mathbf{Z}^{cross}_{i}$ are then processed by a Morph Wrapper (Section~\ref{sec:morph_wrapper}), which warps the primary encoder features $\tilde{\mathbf{F}}^{(1)}_{i}$ using displacement fields and attention maps derived from the cross-modal flow:
\begin{equation}
\mathbf{E}^{cross}_{i} = \textit{MorphWrapper}\big(\tilde{\mathbf{F}}^{(1)}_{i}, \mathbf{Z}^{cross}_{i}\big),
\end{equation}
where $\mathbf{E}^{cross}_{i}$ represents the cross-aligned embeddings. The \textit{cross Morph Wrapper} extracts displacement fields $\mathbf{\Delta}^{cross}_{i}$ and attention maps $\mathbf{A}^{cross}_{i}$ from $\mathbf{Z}^{cross}_{i}$, performing adaptive spatial warping that aligns multi-modal features through learned geometric transformations.
Following cross-modal alignment, the network performs self-refinement to enhance intra-modal consistency. A $MSFE_{self}$ processes the concatenation of the original primary features and the cross-aligned embeddings:
\begin{equation}
\mathbf{Z}^{self}_{i} = \textit{MSFE}_{self}\Big(\text{Concat}\big(\tilde{\mathbf{F}}^{(1)}_{i}, \mathbf{E}^{cross}_{i}\big)\Big),
\end{equation}
where $\mathbf{Z}^{self}_{i}$ captures self-consistent flow patterns that refine the representation while preserving structure. 
Unlike the cross-modal stage, the self Morph Wrapper operates exclusively on the self-refined features, using $\mathbf{Z}^{self}_{i}$ as both the embedding source and the flow guidance:
\begin{equation}
\mathbf{E}^{self}_{i} = \textit{MorphWrapper}\big(\mathbf{Z}^{self}_{i}, \mathbf{Z}^{self}_{i}\big),
\end{equation}
where $\mathbf{E}^{self}_{i}$ represents the self-aligned embeddings.  
The \textit{self Morph Wrapper} similarly extracts displacement fields $\mathbf{\Delta}^{self}_{i}$ and attention maps $\mathbf{A}^{self}_{i}$ from $\mathbf{Z}^{self}_{i}$, performing adaptive warping that preserves structural integrity while refining local spatial details.  
This self-referential deformation enables fine-grained correction and enhancement without interference from cross-modal signals.
The outputs from both cross-modal and self-refinement pathways are then fused through a refinement MSFE that integrates complementary information:
\begin{equation}
\mathbf{Z}^{fused}_{i} = \textit{MSFE}_{refine}\Big(\textit{Concat}\big(\mathbf{E}^{cross}_{i}, \mathbf{E}^{self}_{i}\big)\Big),
\end{equation}
which produces unified flow-estimated embeddings $\mathbf{Z}^{fused}_{i}$ that combine cross-modal alignment with self-consistent refinement. These fused features capture both inter-modal correspondences and intra-modal structural coherence. Unlike the T2I-Adapter, to ensure semantic guidance to the adapter with the text prompt, the fused embeddings are conditioned through cross-attention with text embeddings from the frozen CLIP text encoder $\mathcal{C}_T$: 
\begin{equation}
\mathbf{F}^{final}_{i} = \textit{CrossAttention}\big(\mathbf{Z}^{fused}_{i}, \mathcal{C}_T(c)\big),
\label{eq:final}
\end{equation}
where $c$ denotes the text prompt and $\mathbf{F}^{final}_{i}$ represents the semantically aligned feature embedding at scale $i$. This cross-attention mechanism ensures that the visual conditioning remains consistent with the textual intent throughout the generation process.

\subsubsection{Multi-Scale Flow Estimator (MSFE)}\label{sec:msfe}

\textit{Multi-Scale Flow Estimator (MSFE)} serves to learn hierarchical feature transformations that estimate the Morpho Fields and attention weights for flow-aware guidance across spatial scales. It consists of a lightweight convolutional hierarchy that progressively refines latent representations through multi-resolution filtering and non-linear activation to produce morphological displacement fields and their corresponding attention maps. This lightweight architecture enables rapid convergence during two-stage training while maintaining sufficient capacity for multi-scale flow patterns.
Given an input feature map 
\(\mathbf{Z}_{in} \in \mathbb{R}^{C_{z} \times H_{w} \times W_{w}}\),
the MSFE produces an output \(\mathbf{Z}_{out} \in \mathbb{R}^{(3M) \times H_{w} \times W_{w}}\) through a sequence of convolutional and activation layers:
\begin{equation}
\begin{aligned}
     \mathbf{Z}_{1} = \mathit{LReLU}\big(\mathit{Conv}^{3 \times 3}_{C_{z} \rightarrow C_{1}}(\mathbf{Z}_{in})\big), \quad
    & \mathbf{Z}_{2} = \mathit{LReLU}\big(\mathit{Conv}^{3 \times 3}_{C_{1} \rightarrow C_{2}}(\mathbf{Z}_{1})\big), \\
    \mathbf{Z}_{3} = \mathit{LReLU}\big(\mathit{Conv}^{3 \times 3}_{C_{2} \rightarrow C_{3}}(\mathbf{Z}_{2})\big), \quad
    & \mathbf{Z}_{out} = \mathit{Conv}^{3 \times 3}_{C_{3} \rightarrow 3M}(\mathbf{Z}_{3}),
\end{aligned}
\end{equation}
where $\mathit{LReLU}$ denotes the Leaky ReLU activation with a negative slope of $0.1$, and $M$ is the number of sampling points.

The output \(\mathbf{Z}_{out}\) is then decomposed through channel splitting to extract the Morpho Fields and attention weights:
\begin{equation}
\mathbf{\Delta} = \mathbf{Z}_{out}[0:2M, :, :], \quad \mathbf{A} = \mathbf{Z}_{out}[2M:3M, :, :],
\end{equation}
where the first \(2M\) channels encode the Morpho Fields \(\mathbf{\Delta} \in \mathbb{R}^{M \times 2 \times H_{w} \times W_{w}}\) defining morphological spatial offsets (with 2 coordinates per sampling point), and the remaining \(M\) channels represent the morphological attention weights \(\mathbf{A} \in \mathbb{R}^{M \times H_{w} \times W_{w}}\). The combined representation \([\mathbf{\Delta}, \mathbf{A}] = \mathbf{Z}_{out}\) directly serves as the conditioning feature for the \textit{MorphWrapper} function, enabling end-to-end learning of morphological transformations.

\subsubsection{Morph Wrapper}\label{sec:morph_wrapper}

As illustrated in Fig.~\ref{fig:wrap}, the \textit{Morph Wrapper} operates as a morphological attention mechanism that unifies spatial embeddings and attention maps through adaptive sampling. 
Unlike conventional deformable attention with fixed offset patterns, the Morph Wrapper dynamically wraps features, aligning multi-scale embeddings to context-dependent spatial configurations via learned attention modulation to preserve spatial information that is typically lost in standard cross-attention.
The \textit{MorphWrapper} function takes two inputs:
a source embedding 
\(\mathbf{P} \in \mathbb{R}^{C_{p} \times H_{w} \times W_{w}}\)
and the MSFE output 
\(\mathbf{Z} \in \mathbb{R}^{(3M) \times H_{w} \times W_{w}}\),
where \(M\) denotes the number of sampling points. \(\mathbf{Z}\) contains both Morpho Fields and attention maps 
$[\mathbf{\Delta}, \mathbf{A}] = \mathbf{Z},$
with \(\mathbf{\Delta} \in \mathbb{R}^{M \times 2 \times H_{w} \times W_{w}}\) representing morphological spatial offsets that define adaptive sampling locations, and 
\(\mathbf{A} \in \mathbb{R}^{M \times H_{w} \times W_{w}}\) denoting attention weights.
The mechanism synthesizes wrapped features via differentiable bilinear interpolation:
\begin{equation}
\mathbf{P}'_{m} = \mathit{GridSample}\big(\mathbf{P}, \mathbf{\Delta}_{m}\big),
\quad m = 1, \ldots, M.
\end{equation}
Each warped embedding is modulated by its attention map through softmax-normalized aggregation:
\begin{equation}
\footnotesize
\textit{MorphWrapper}(\mathbf{P}, \mathbf{Z})(p) =
\sum_{m=1}^{M}
\frac{\exp(\mathbf{A}_{m}(p))}{\sum_{j=1}^{M} \exp(\mathbf{A}_{j}(p))}
\, \mathbf{P}'_{m}(p),
\end{equation}
where \(p\) indexes spatial positions.
This differentiable morphological attention mechanism enables flexible geometric adaptation, enhancing correspondence between conditional inputs and latent representations while maintaining end-to-end trainability.

\subsection{Fusion with Backbone}

At each of the $L$ scales, adapter outputs the final condition-aware features $\{\mathbf{F}_{i}^{final}\}_{i=1}^{L}$ (Eq.~\ref{eq:final}). These features are integrated with the corresponding encoder stages of the frozen Stable Diffusion 1.5 UNet through point-wise addition:
\begin{equation}
\mathbf{U}_{i}^{final} = \mathbf{U}_{i}^{enc} + \mathbf{F}_{i}^{final},
\end{equation}
where $\mathbf{U}_{i}^{enc}$ denotes the intermediate feature map from the UNet encoder at stage $i$, and $\mathbf{U}_{i}^{final}$ represents the fused encoder representation after adapter injection.  
This point-wise additive fusion retains the UNet’s pretrained representational strength while infusing globally consistent, structure-aware conditioning from the Adapter.  
Through staged flow estimation from the \textit{Morpho Fields}, self and cross refinements through \textit{MSFE}, and text-conditioned guidance, the adapter progressively enhances spatial coherence and semantic alignment across scales, ensuring efficient and stable conditioning throughout the denoising process.

\section{Experiments}

\paragraph{Datasets:} We train \our{} on MS-COCO 2017~\cite{mscoco2017} and MultiGen-20M~\cite{multigen20m} datasets. MS-COCO provides diverse real-world images with rich annotations across multiple object categories, while MultiGen-20M offers large-scale multi-modal conditioning pairs essential for learning universal cross-modal representations. For evaluation, we utilize the MS-COCO 2017 validation set comprising 5,000 images with 25,000 image-caption pairs, enabling assessment of generation quality and fidelity.

\subsection{Results and Discussion}
\begin{table*}[t]
\centering
\small
\caption{
Comparison of FID (↓) and CLIP (↑) scores across four conditions. 
Model complexity includes Parameters (M), FLOPs (G), and Memory (GB), where Memory reflects total training memory to load the SD1.5 with the adapter.  
Best are in \textbf{bold}.
}

\label{tab:merged_fid_clip}
\resizebox{\textwidth}{!}{%
\begin{tabular}{
    l|c|c|c|
    cc|cc|cc|cc
}
\hline
\multirow{2}{*}{\textbf{Model}} &
\multirow{2}{*}{\textbf{Params. (M)}} &
\multirow{2}{*}{\textbf{FLOPs (G)}} &
\multirow{2}{*}{\textbf{Memory (GB)}} &
\multicolumn{2}{c|}{\textbf{Canny}} &
\multicolumn{2}{c|}{\textbf{Depth}} &
\multicolumn{2}{c|}{\textbf{Sketch}} &
\multicolumn{2}{c}{\textbf{Segmentation}} \\ 
\cline{5-12}
& & & &
\textbf{FID~$\downarrow$} & \textbf{CLIP~$\uparrow$} &
\textbf{FID~$\downarrow$} & \textbf{CLIP~$\uparrow$} &
\textbf{FID~$\downarrow$} & \textbf{CLIP~$\uparrow$} &
\textbf{FID~$\downarrow$} & \textbf{CLIP~$\uparrow$} \\ 
\hline

ControlNet~\cite{controlnet} & 361.28 & 116.61 & 24$\times$2 &
22.84 & 27.41 &
25.68 & 27.51 &
24.93 & 27.38 &
27.06 & 27.04 \\

T2I-Adapter~\cite{t2i} & \underline{77.37} & \underline{29.97} & 24$\times$1 &
23.73 & 26.72 &
26.03 & 26.46 &
26.51 & 27.21 &
27.66 & 26.99 \\

ControlNet++~\cite{controlnet++} & 361.28 & 116.61 & 24$\times$2 &
23.59 & 27.09 &
25.19 & 27.56 &
24.71 & 27.15 &
26.23 & 27.11 \\

Uni-ControlNet~\cite{unicontrolnet} & 1271.42 & 210.77 & 24$\times$8 &
23.11 & 27.21 &
24.92 & 27.43 &
24.56 & 27.54 &
25.33 & 27.49 \\

CtrLoRA~\cite{ctrlora} & 398.28 & 135.15 & 24$\times$4 &
22.59 & 27.16 &
25.65 & 26.34 &
26.02 & 25.46 &
25.73 & 25.62 \\

UniCon~\cite{unicon} & 150.00 & 111.62 & 24$\times$8 &
22.86 & 26.92 &
25.30 & 27.42 &
24.61 & 27.13 &
26.42 & 27.04 \\

\hline
\textbf{\our$_{Ind}$} & 365.25 & 135.82 & 24$\times$1 &
22.37 & 28.09 &
24.12 & 27.90 &
24.38 & 27.76 &
25.18 & 27.89 \\

\textbf{\our$_{Pre}$} & 365.25 & 135.82 & 24$\times$1 &
\textbf{21.48} & \textbf{28.52} &
\textbf{22.44} & \textbf{28.18} &
\textbf{23.21} & \textbf{28.54} &
\textbf{23.91} & \textbf{27.91} \\

\hline
\end{tabular}%
}
\end{table*}

\begin{table}[t]
\centering
\scriptsize
\setlength{\tabcolsep}{1pt}
\caption{Comparison of FID (↓) and CLIP (↑) scores across four conditions. 
Model complexity includes Parameters (M), and FLOPs (G) for each adapter with the SDXL backbone.  
Best are in \textbf{bold}.}
\label{tab:sdxl_fid_clip}
\resizebox{\linewidth}{!}{
\begin{tabular}{l|c|c|cc|cc|cc|cc}
\hline
\multirow{2}{*}{\textbf{Model}} &
\multirow{2}{*}{\textbf{Params (M)}} &
\multirow{2}{*}{\textbf{FLOPs (G)}} &
\multicolumn{2}{c|}{\textbf{Canny}} &
\multicolumn{2}{c|}{\textbf{Depth}} &
\multicolumn{2}{c|}{\textbf{Sketch}} &
\multicolumn{2}{c}{\textbf{Segmentation}} \\
\cline{4-11}
& & &
\textbf{FID~$\downarrow$} & \textbf{CLIP~$\uparrow$} &
\textbf{FID~$\downarrow$} & \textbf{CLIP~$\uparrow$} &
\textbf{FID~$\downarrow$} & \textbf{CLIP~$\uparrow$} &
\textbf{FID~$\downarrow$} & \textbf{CLIP~$\uparrow$} \\
\hline

ControlNet~\cite{controlnet} &
1250.98  & 1336.69 &
23.75 & 31.71 &
26.42 & 31.68 &
25.87 & 31.64 &
27.82 & 31.58 \\

T2I-Adapter~\cite{t2i} &
79.03 & 29.95 &
24.89 & 31.42 &
27.18 & 31.28 &
27.64 & 31.36 &
28.91 & 31.19 \\

ControlNet++~\cite{controlnet++} &
1250.98 & 1336.69 &
24.12 & 31.58 &
26.09 & 31.54 &
25.43 & 31.61 &
27.35 & 31.52 \\

\hline
\textbf{\our$_{Ind}$} &
384.83 & 140.17 &
21.18 & 31.86 &
23.42 & 31.73 &
23.89 & 31.92 &
24.54 & 31.68 \\

\textbf{\our$_{Pre}$} &
384.83 & 140.17 &
\textbf{20.35} & \textbf{32.18} &
\textbf{21.26} & \textbf{31.89} &
\textbf{21.98} & \textbf{32.21} &
\textbf{22.67} & \textbf{31.76} \\

\hline
\end{tabular}
}
\end{table}
\begin{figure*}[t]
    \centering
    \includegraphics[width=1\linewidth]{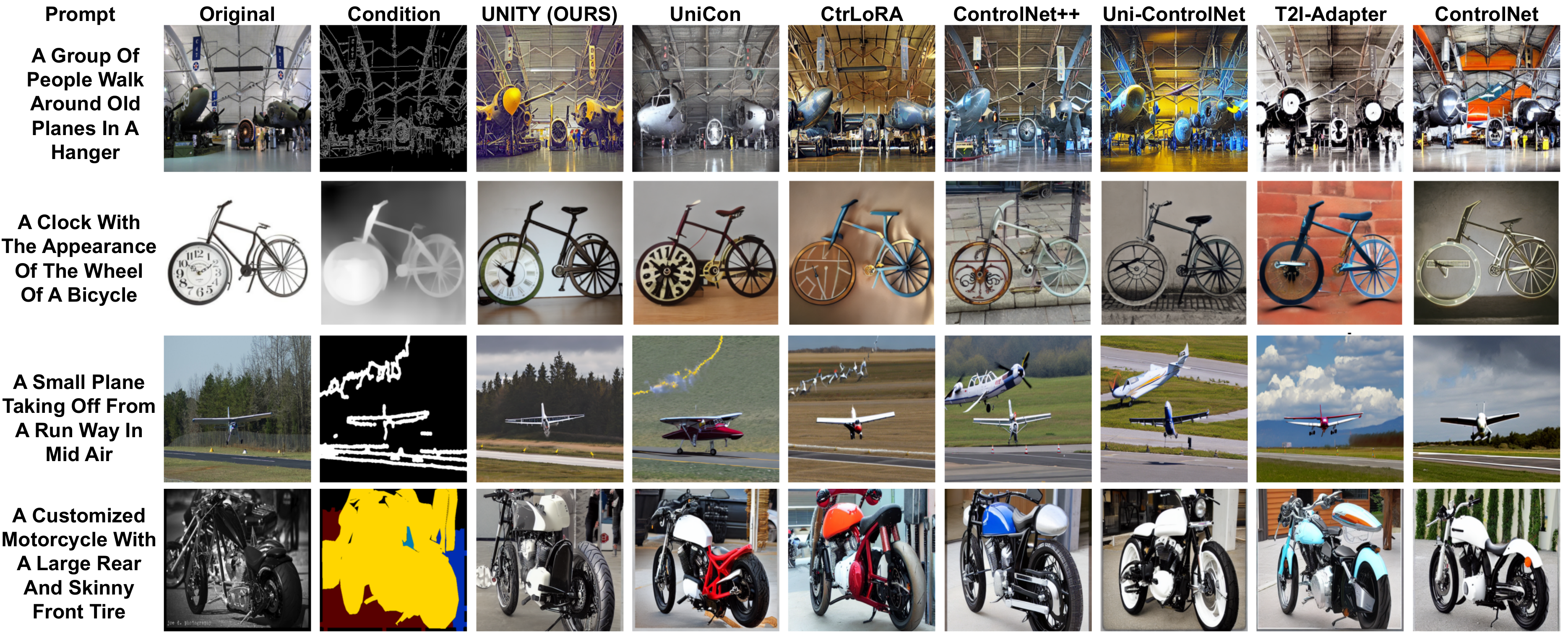}
    \caption{Qualitative results across conditions shows that \our{} consistently produces the most accurate and semantically aligned outputs.}
    \label{fig:qsamples}
\end{figure*}

Table~\ref{tab:merged_fid_clip} presents a comprehensive comparison of \our-Adapter against SoTA conditional image generation methods across four conditioning modalities. We analyze model complexity, quantitative performance, and the effectiveness of our training paradigm.

\paragraph{Quantitative Performance Discussion.}
As shown in Table~\ref{tab:merged_fid_clip}, \our$_{\textit{Pre}}$ consistently achieves the best FID and CLIP scores across all four conditions on SD1.5, reaching an average FID of 22.76 and CLIP of 28.29. This corresponds to a 7.03\% improvement in FID and 3.17\% in CLIP over the next-best baseline, Uni-ControlNet (24.48 FID, 27.42 CLIP). The largest gains occur on the Depth condition, where FID decreases from 24.92 to 22.44 (9.95\%) and CLIP increases to 28.18, surpassing ControlNet++ (27.56). Consistent improvements are also observed for Canny (21.48 FID, 28.52 CLIP) and Sketch (23.21 FID), indicating stronger structural fidelity and semantic alignment across diverse control modalities. Notably, these gains are achieved with substantially fewer parameters and lower FLOPs than Uni-ControlNet, highlighting the efficiency of our approach.

Table~\ref{tab:sdxl_fid_clip} further confirms this trend on SDXL, where \our$_{\textit{Pre}}$ achieves the lowest FID across all conditions—20.35 (Canny), 21.26 (Depth), 21.98 (Sketch), and 22.67 (Segmentation)—while also obtaining the highest CLIP scores, up to 32.21. Compared with ControlNet, the FID improvement on Canny alone reaches 14.32\%, decreasing from 23.75 to 20.35, with similarly consistent gains across other conditions. Moreover, the universal pretraining stage yields a 5.22\% FID improvement over \our$_{\textit{Ind}}$ on SD1.5, with comparable margins on SDXL, including a 4.91\% gain on Depth. Overall, these results demonstrate that joint cross-condition pretraining improves structural consistency and semantic robustness without introducing additional inference-time overhead.

\paragraph{Qualitative Analysis:} Figure~\ref{fig:qsamples} compares \our{} against eight baseline models across four diverse prompts, testing attribute fidelity and concept fusion. In \textit{Row 1}, \our{} generates realistic hangar scenes with accurate lighting and spatial composition, while baselines suffer severe color degradation: UniCon produces oversaturated blue-yellow tones; CtrLoRA introduces heavy yellow-orange casts, ControlNet++ displays unnatural blue tinting; Uni-ControlNet generates excessive yellow coloring; T2I-Adapter creates harsh contrast, destroying detail; and ControlNet exhibits orange-red contamination, eliminating photorealism. \textit{Row 2} demonstrates \our's superior concept fusion, integrating clock faces within bicycle wheels with clear visibility, whereas UniCon produces abstract forms, CtrLoRA generates poor detail clarity, ControlNet++ creates blurry outputs, Uni-ControlNet obscures features, T2I-Adapter adds distracting backgrounds, and ControlNet compromises legibility. 

\textit{Row 3} shows \our{} maintaining photorealistic aircraft quality while baselines produce overexposure (UniCon), invalid coloring (CtrLoRA, Uni-ControlNet, T2I-Adapter), a toy-like appearance (ControlNet++), or pure silhouettes (ControlNet). \textit{Row 4} highlights \our's accurate motorcycle customization rendering versus baselines' abstract renderings, poor detail, excessive coloring, or texture destruction. \our{} consistently maintains photorealism, accurate colors, and compositional coherence while baselines systematically fail through color contamination and detail loss.

\paragraph{Model Complexity Analysis:}
As summarized in Tables~\ref{tab:merged_fid_clip} and~\ref{tab:sdxl_fid_clip}, \our{} achieves SoTA FID and CLIP scores while maintaining competitive computational cost. On SD1.5, it uses 365.25M parameters and 135.82G FLOPs, comparable to ControlNet (361.28M, 116.61G) but requiring only $24\times1$ memory, whereas ControlNet, ControlNet++, and CtrLoRA demand $24\times2$--$24\times4$, and Uni-ControlNet scales to $24\times8$. Despite a 4.72$\times$ parameter reduction relative to Uni-ControlNet (1271.42M, 210.77G), \our$_{\textit{Pre}}$ achieves lower FID in all conditions, demonstrating superior performance-efficiency trade-offs.

On SDXL, the advantage becomes more pronounced. \our{} requires only 384.83M parameters and 140.17G FLOPs, drastically lower than ControlNet and ControlNet++ (1250.98M, 1336.69G), yet improves Canny FID from 23.75 to 20.35 and raises CLIP from 31.71 to 32.18. Compared to T2I-Adapter (79.03M, 29.95G), \our{} uses moderately higher compute but consistently achieves substantially better FID (e.g., 21.26 vs.\ 27.18 on Depth). With constant parameter size across $K$ conditions and shared universal pretraining, the framework reduces multi-task training cost by 37.5\% while preserving top-tier generative quality under a single-condition memory budget.

\section{Ablation Study}\label{sec:ablation}
\begin{table}[!thb]
\centering
\caption{Ablation (Case 1): No-Text-Prompt Settings. \textit{Without Prompt to Adapter (w/o P2A)} and \textit{Without Prompt to Stable Diffusion and Adapter (w/o P2SD+A)}.}
\label{tab:no_prompts}
\renewcommand{\arraystretch}{1.15}
\setlength{\tabcolsep}{4pt}
\resizebox{\linewidth}{!}{
\begin{tabular}{l|cc|cc|cc|cc}
\toprule
\multirow{2}{*}{\textbf{Model}} 
& \multicolumn{4}{c|}{\textbf{w/o P2A}} 
& \multicolumn{4}{c}{\textbf{w/o P2SD+A}} \\
\cmidrule(lr){2-5} \cmidrule(lr){6-9}
& \multicolumn{2}{c|}{\textbf{Sketch}} 
& \multicolumn{2}{c|}{\textbf{Segmentation}} 
& \multicolumn{2}{c|}{\textbf{Sketch}} 
& \multicolumn{2}{c}{\textbf{Segmentation}} \\
\cmidrule(lr){2-3} \cmidrule(lr){4-5}
\cmidrule(lr){6-7} \cmidrule(lr){8-9}
& \textbf{FID $\downarrow$} & \textbf{CLIP $\uparrow$} 
& \textbf{FID $\downarrow$} & \textbf{CLIP $\uparrow$}
& \textbf{FID $\downarrow$} & \textbf{CLIP $\uparrow$} 
& \textbf{FID $\downarrow$} & \textbf{CLIP $\uparrow$} \\
\midrule
ControlNet~\cite{controlnet} 
& 32.21 & 19.23 & 31.62 & 19.30 
& 38.84 & 16.98 & 37.72 & 15.97 \\

T2I-Adapter~\cite{t2i}
& -- & -- & -- & -- 
& 36.59 & 17.53 & 35.50 & 16.45 \\

ControlNet++~\cite{controlnet++}
& 32.28 & 19.26 & 31.68 & 19.33 
& 38.90 & 17.02 & 37.81 & 16.01 \\

Uni-ControlNet~\cite{unicontrolnet} 
& 30.82 & 20.16 & 30.07 & 20.20 
& 36.28 & 16.87 & 35.44 & 16.47 \\

CtrLoRA~\cite{ctrlora}
& 30.12 & 20.20 & 29.89 & 20.24 
& 35.34 & 16.91 & 34.52 & 16.50 \\

UniCon~\cite{unicon}
& 30.91 & 19.76 & 31.12 & 19.49 
& 37.10 & 17.61 & 35.43 & 17.97 \\

\midrule
\textbf{UNITY}$_{Pre}$
& \textbf{24.94} & \textbf{27.58} & \textbf{25.80} & \textbf{27.04}
& \textbf{28.83} & \textbf{24.16} & \textbf{30.08} & \textbf{23.20} \\

\bottomrule
\end{tabular}
}

\bigskip 
\footnotesize
\caption{
Comprehensive ablation study of \our-Adapter. 
Three experimental groups analyze: (1) text-conditioning removal, (2) architectural design components, and (3) multi-condition pretraining strategies. 
FID ($\downarrow$) and CLIP ($\uparrow$) scores are reported for Sketch and Segmentation conditions.
}
\label{tab:unified_ablation}
\setlength{\tabcolsep}{3pt}
\renewcommand{\arraystretch}{0.95}
\resizebox{\linewidth}{!}{%
\begin{tabular}{l|cc|cc}
\toprule
\multirow{2}{*}{\hspace{2.6cm}\textbf{Cases}} & 
\multicolumn{2}{c|}{\textbf{Sketch}} & 
\multicolumn{2}{c}{\textbf{Segmentation}} \\ 
\cmidrule(lr){2-3}\cmidrule(lr){4-5}
 & FID~$\downarrow$ & CLIP~$\uparrow$ & FID~$\downarrow$ & CLIP~$\uparrow$ \\ 
\midrule

\textbf{\textit{Case 2:} Architectural Design Variants} & & & & \\
Cross-modal Flow w/ Cross-Attention    & 25.35 & 27.31 & 26.29 & 26.76 \\
Self-refinement w/ Self-Attention      & 25.88 & 27.13 & 26.75 & 26.54 \\
w/o Text Cross-Attention               & 24.73 & 26.65 & 25.65 & 25.98 \\
Single Spatial Encoder                 & 26.40 & 26.92 & 27.38 & 26.33 \\
\midrule

\textbf{\textit{Case 3:} Multi-Condition Pretraining} & & & & \\
Canny + Depth                          & 24.79 & 27.48 & 25.63 & 26.93 \\
Depth + Segmentation                   & 24.64 & 27.63 & 25.45 & 27.07 \\
Depth + Sketch + Segmentation          & 23.94 & 28.12 & 24.77 & 27.53 \\
Canny + Sketch + Segmentation          & 23.88 & 28.17 & 24.70 & 27.58 \\
\midrule

\textbf{\our}$_{Pre}$ & \textbf{23.21} & \textbf{28.54} & \textbf{23.91} & \textbf{27.91} \\
\bottomrule
\end{tabular}
}

\end{table}

In Table~\ref{tab:no_prompts} and~\ref{tab:unified_ablation}, we systematically validate each component of \our-Adapter through targeted ablations spanning text conditioning, architectural design, and multi-condition pretraining. 
All variants use identical training settings and are evaluated on the MS-COCO validation set with Sketch and Segmentation conditioning, showing that every proposed component contributes meaningfully.

\paragraph{\textit{Case 1 -} No-Text-Prompt Settings: } 
We evaluate two challenging ablation scenarios that progressively suppress text conditioning: removing prompts from the adapter only (\textit{w/o P2A}), and removing prompts from both Stable Diffusion and the adapter (\textit{w/o P2SD+A}). As reported in Table~\ref{tab:no_prompts}, \textbf{UNITY}$_{Pre}$ consistently dominates all baselines across both structural modalities under these constrained settings. In the \textit{w/o P2A} regime, \textbf{UNITY}$_{Pre}$ achieves substantial improvements, reducing FID by 5--7 points over strong competitors such as CtrLoRA and Uni-ControlNet, while simultaneously boosting CLIP scores by $+$7 to $+$8 points, reflecting markedly superior semantic consistency even in the absence of adapter-level text conditioning. When prompts are entirely withheld from both Stable Diffusion and the adapter (\textit{w/o P2SD+A}), competing baselines degrade sharply, with FID increasing by $\sim$6--8 points and CLIP scores declining by $\sim$3 points, unambiguously evidencing their heavy reliance on explicit textual supervision. In stark contrast, \textbf{UNITY}$_{Pre}$ preserves consistently superior generation fidelity (lowest FID: 28.83 / 30.08) and semantic alignment (highest CLIP: 24.16 / 23.20) across both Sketch and Segmentation conditions, demonstrating robust structure-driven controllability that does not collapse in text-free inference regimes. These results highlight a fundamental advantage of our unified pretraining strategy: the model internalizes rich structural priors that sustain high-quality, semantically coherent synthesis independent of text prompt availability.

\paragraph{\textit{Case 2 - } Architectural Design Variants: }
In Table~\ref{tab:unified_ablation}, substituting MAF components with standard attention mechanisms validates our flow-based design. \textit{Cross-modal Flow w/ Cross-Attention} replaces flow estimation with vanilla cross-attention, degrading Sketch FID by 9.2\% (25.35) and CLIP by 4.3\% (27.31), demonstrating that explicit spatial modeling outperforms attention-only alignment. \textit{Self-refinement w/ Self-Attention} yields worse performance (FID 25.88, CLIP 27.13), confirming that morphological warping's geometric adaptivity exceeds self-attention's global relationship modeling. Removing text cross-attention from the final fusion stage (\textit{w/o Text Cross-Attention}) causes the largest CLIP drop of 1.89 points (6.6\%), proving that text-guided fusion is essential for semantic alignment. \textit{Single Spatial Encoder} shows the highest FID degradation (+3.19, +13.7\%), validating that dual-pathway encoding enables richer cross-modal representations than sequential processing.

\paragraph{\textit{Case 3 - } Multi-Condition Pretraining Strategies: }
Progressive improvement with modality diversity validates our universal pretraining paradigm, as can be observed in Table~\ref{tab:unified_ablation}. Two-modality configurations (\textit{Canny + Depth}, \textit{Depth + Segmentation}) average FID 1.50 above baseline and CLIP 0.98 below, establishing dual-condition baselines. Three-modality combinations significantly improve: \textit{Depth + Sketch + Segmentation} achieves FID 23.94 (+0.73, +3.1\%) and CLIP 28.12 (-0.42, -1.5\%), while \textit{Canny + Sketch + Segmentation} reaches FID 23.88 and CLIP 28.17. The consistent trend shows that increased modality diversity during universal pretraining establishes stronger shared representations. Our full four-modality pretraining (Canny, Depth, Sketch, Segmentation) achieves optimal performance (FID 23.21, CLIP 28.54), demonstrating that maximal conditioning diversity enables the most robust structure-aware semantic learning. This validates the universal-to-specialized training paradigm's ability to leverage cross-modal information effectively.

\section{Conclusion}

In this work, we presented \our, a novel adapter that fundamentally addresses scalability and efficiency limitations in multi-modal conditional diffusion models. Our two-stage training paradigm enables significant training reduction while maintaining constant parameters regardless of conditioning diversity. The proposed novel Morphable Attention Flow Network with Dual Spatial Encoders, Multi-Scale Flow Estimators, and Morph Wrapper modules provides precise geometric alignment through learned displacement fields and attention-based fusion to preserve spatial information. Unlike existing methods that scale linearly in training cost and require separate adapters per modality, \our-Adapter offers global guidance without redundant parallel denoising, substantially reducing computational overhead while supporting flexible single or composite conditioning at deployment. This work presents an efficient paradigm for controllable generation, showing that principled architecture and strategic training can effectively bridge generalization and specialization.


\bibliographystyle{splncs04}
\bibliography{main}

\clearpage

\begin{center}
    {\LARGE \textbf{Supplementary Material}}
\end{center}
\vspace{1em}

\setcounter{section}{0}

\section{Training and Evaluation}

We evaluate two configurations: \text{\our$_{\textit{Ind}}$} one that independently trains adapters without the universal-to-specialized transfer, and \text{\our$_{\textit{Pre}}$}, which employs the proposed two-stage strategy in Section~\ref{sec:framework}, which are trained for 100k steps. 
In the \textit{Universal Stage}, the model is trained on the MS-COCO training set using all four conditioning modalities (Canny, Depth, Sketch, and Segmentation) to train jointly. During the \textit{Specialization Stage}, each is individually further finetuned on both MS-COCO and MultiGen-20M combined, while the Segmentation adapter is refined solely on MS-COCO due to its annotations.
We assess generation performance using Fréchet Inception Distance (FID)~\cite{fid} for visual fidelity and CLIP Score~\cite{clipscore} for text-image alignment. All experiments use a frozen Stable Diffusion v1.5 backbone for fair comparison. We also trained baselines on identical datasets for $100k$ steps. Optimization is performed with AdamW~\cite{adamw} using a constant learning rate \(5 \times 10^{-6}\), \(\beta_1 = 0.9\), \(\beta_2 = 0.999\), \(\epsilon = 1 \times 10^{-8}\), and weight decay of \(1 \times 10^{-2}\). A linear warm-up of 500 steps stabilizes early training before transitioning to constant learning rate scheduling. \textit{Further experiments on additional conditioning types are provided in the supplementary material.}

\section{Additional Experiments}

\begin{table}[t]
\centering
\small
\caption{Quantitative comparison on additional conditioning modalities. 
FID$\downarrow$ and CLIP$\uparrow$ scores are reported. Best results in \textbf{bold}.}
\label{tab:additional_exp}

\begin{tabularx}{\linewidth}{l|XXXXXXXX}
\toprule
\multirow{2}{*}{\textbf{Model}} &
\multicolumn{2}{c}{\textbf{Normal Map}} &
\multicolumn{2}{c}{\textbf{Openpose}} &
\multicolumn{2}{c}{\textbf{Color Palette}} &
\multicolumn{2}{c}{\textbf{Dehaze}} \\
\cmidrule(lr){2-3} \cmidrule(lr){4-5} \cmidrule(lr){6-7} \cmidrule(lr){8-9}
 & FID$\downarrow$ & CLIP$\uparrow$
 & FID$\downarrow$ & CLIP$\uparrow$
 & FID$\downarrow$ & CLIP$\uparrow$
 & FID$\downarrow$ & CLIP$\uparrow$ \\
\midrule
ControlNet~\cite{controlnet}     & 24.46 & 27.22 & 25.70 & 27.08 & 26.79 & 26.89 & 26.12 & 26.96 \\
Uni-ControlNet~\cite{unicontrolnet} & 24.29 & 27.39 & 25.08 & 27.33 & 26.33 & 27.18 & 25.86 & 27.22 \\
CtrLoRA~\cite{ctrlora}        & 24.94 & 26.78 & 26.14 & 26.55 & 26.95 & 26.33 & 26.69 & 26.49 \\
\midrule
\our{}$_{Ind}$ & 23.85 & 27.96 & 24.92 & 27.39 & 25.91 & 27.22 & 25.01 & 27.16 \\
\our{}$_{Pre}$ & \textbf{22.08} & \textbf{28.72}
               & \textbf{23.78} & \textbf{28.36}
               & \textbf{24.51} & \textbf{27.94}
               & \textbf{24.18} & \textbf{27.82} \\
\bottomrule
\end{tabularx}
\end{table}

\subsection{SoTA Results on Stable Diffusion 1.5 with Additional Conditions}

We conducted further experiments to validate the generalization capability and robustness of \our{} in various conditioning modalities. We extend our evaluation to four additional conditioning types: \textbf{Normal Map}, \textbf{Openpose}, \textbf{Color Palette}, and \textbf{Dehaze}. Following the two-stage training paradigm established in Section~\ref{sec:framework}, we train adapters from scratch using the Universal-to-Specialized strategy on these four new conditions and compare against the three top-performing baselines from main experiments: \textbf{Uni-ControlNet}~\cite{unicontrolnet}, \textbf{CtrLoRA}~\cite{ctrlora}, and \textbf{ControlNet}~\cite{controlnet}. Unlike the main experiments which employed Canny, Depth, Sketch, and Segmentation, this configuration trains exclusively on the new conditioning quartet to assess \our{}'s adaptability to complementary signal types spanning geometric surface properties (Normal Map), human pose articulation (Openpose), content-preserving stylization (Color Palette), and atmospheric degradation modeling (Dehaze). 

\textbf{Conditional Image Generation and Training Configuration:}
We generate all conditioning inputs using standard annotators for normal maps, Openpose, Color Palette, and Dehaze modalities, applied to both MS-COCO 2017 and MultiGen-20M to form paired training data. MS-COCO provides diverse real-world imagery, while MultiGen-20M supplies large-scale synthetic variation, enabling robust cross-modal learning. All conditioning maps are resized to $512 \times 512$ and normalized for compatibility with the Stable Diffusion v1.5 backbone. Training follows a two-stage strategy: a 50k-step Universal Stage that jointly learns across all conditions to build shared representations, followed by a 50k-step Specialization Stage that fine-tunes each condition independently. This preserves the total 100k-step training budget used in the main experiments while maintaining a constant parameter count (365.25M) and FLOPs (135.82G) irrespective of the number of supported conditioning types.

\subsubsection{Results and Discussion}

Table~\ref{tab:additional_exp} presents a detailed quantitative comparison across four additional conditioning modalities. Several clear trends emerge from the FID and CLIP metrics. Among the baselines, Uni-ControlNet consistently performs better than ControlNet and CtrLoRA, although the gains remain modest, with improvements typically in the range of a few tenths of a point for both FID and CLIP, based on the modality. 

Both versions of our method provide consistent and meaningful improvements across all modalities. Even without any pretraining, \our{}$_{Ind}$ already surpasses every baseline, reducing FID by roughly one point or more in several settings and improving CLIP scores by a noticeable margin. These gains reflect the advantages of our unified architecture together with shared conditioning representations.
The best results are achieved by the pretrained variant, \our{}$_{Pre}$, which obtains the strongest metric values in every modality. For Normal Map, it reduces the FID score from 24.29 for Uni ControlNet to 22.08 and increases the CLIP value from 27.39 to 28.72. For Openpose, it improves the FID score from 25.08 to 23.78. For Color Palette, it reduces FID from 26.33 to 24.51. For Dehaze, it improves the FID score from 25.86 to 24.18. In all cases, CLIP similarity also increases. The largest gains appear in structure-focused modalities, confirming that unified pretraining enhances geometric reasoning and the alignment between the condition and the generated image.

The quantitative results show that unified pretraining followed by a specialization stage provides clear benefits across both geometry-oriented and appearance-oriented modalities. The consistent reductions in FID and increases in CLIP demonstrate enhanced semantic fidelity together with reliable controllability in a broad range of conditional image generation.

\end{document}